\documentclass[10pt,twocolumn,letterpaper]{article}

\usepackage{iccv}
\usepackage{times}
\usepackage{epsfig}
\usepackage{graphicx}
\usepackage{amsmath}
\usepackage{amssymb}

\DeclareMathOperator*{\argmin}{arg\,min}
\DeclareMathOperator*{\argmax}{arg\,max}
\usepackage{array}
\usepackage[breaklinks=true,bookmarks=false]{hyperref}

\iccvfinalcopy 

\def\iccvPaperID{} 

\begin{document}

\title{Improving CNN classifiers by estimating test-time priors}

\author{
  Milan Sulc, Jiri Matas \\
  Dept. of Cybernetics, FEE CTU in Prague\\
  Technicka 2, Prague, Czech Republic \\
  {\tt\small sulcmila,matas@fel.cvut.cz}
}

\maketitle

\begin{abstract}
The problem of different training and test set class priors is addressed in the context of CNN classifiers. We compare two different approaches to estimating the new priors: an existing Maximum Likelihood Estimation approach (optimized by an EM algorithm or by projected gradient descend) and a proposed Maximum a Posteriori approach, which increases the stability of the estimate by introducing a Dirichlet hyper-prior on the class prior probabilities.
Experimental results show a significant improvement on the fine-grained classification tasks using known evaluation-time priors, increasing the top-1 accuracy by~4.0\% on the FGVC iNaturalist 2018 validation set and by~3.9\% on the FGVCx Fungi 2018 validation set. Estimation of the unknown test set priors noticeably increases the accuracy on the PlantCLEF dataset, allowing a single CNN model to achieve state-of-the-art results and outperform the competition-winning ensemble of 12 CNNs. The proposed Maximum a Posteriori estimation increases the prediction accuracy by~2.8\% on PlantCLEF 2017 and by 1.8\% on FGVCx Fungi, where the existing MLE method would lead to a decrease accuracy.
\end{abstract}


\section{Introduction}

A common assumption of many machine learning algorithms is that the training set is 
independently sampled from the same data distribution as the test data \cite{bishop2006pattern, goodfellow2016deep, hastie2001elements}.
In practical computer vision tasks, this assumption is often violated - training samples may be obtained from diverse sources where classes appear with frequencies differing from the test-time.
For instance, for the task of fine-grained recognition of plant species from images, training examples can be downloaded from an online encyclopedia. However, the number of photographs of a species in the encyclopedia may not correspond to the species incidence or to the frequency a species is queried in a plant identification service.

\begin{figure}[hbt]
\centering
\includegraphics[height=1.19cm]{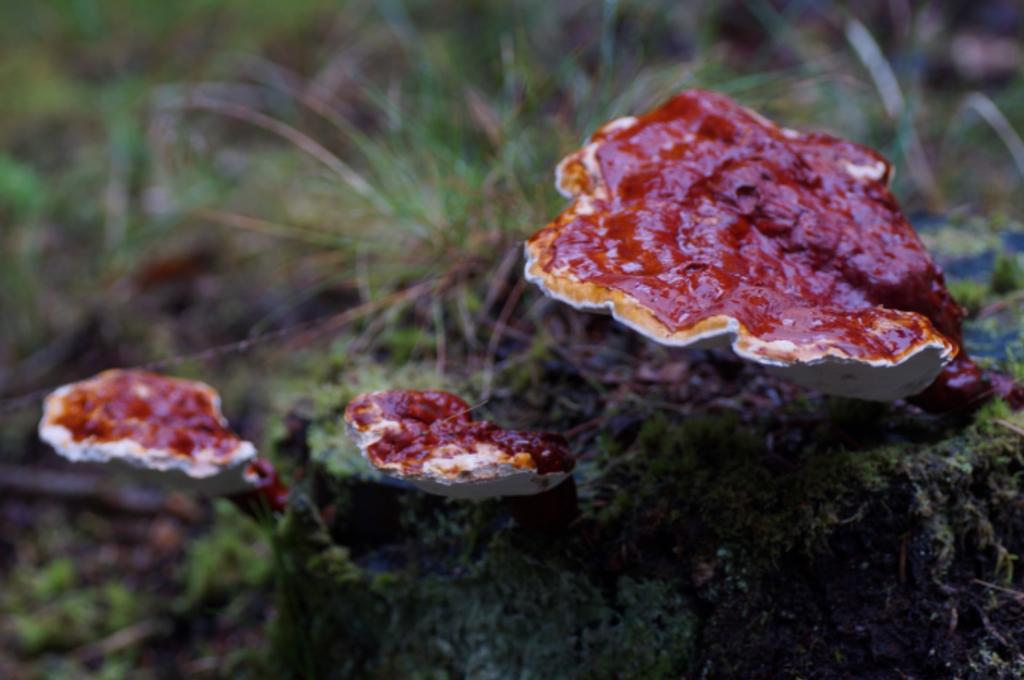}
\includegraphics[height=1.19cm]{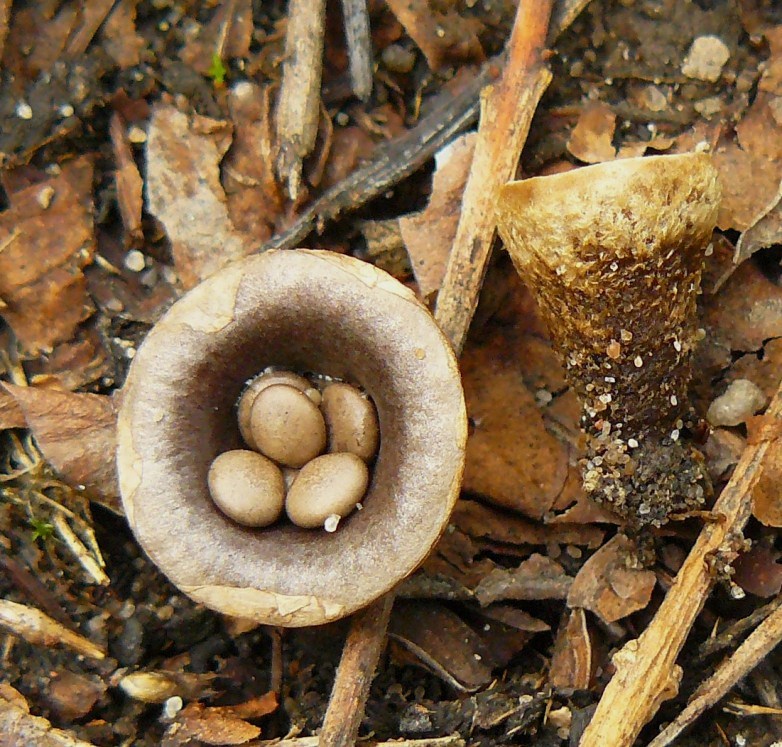}
\includegraphics[height=1.19cm]{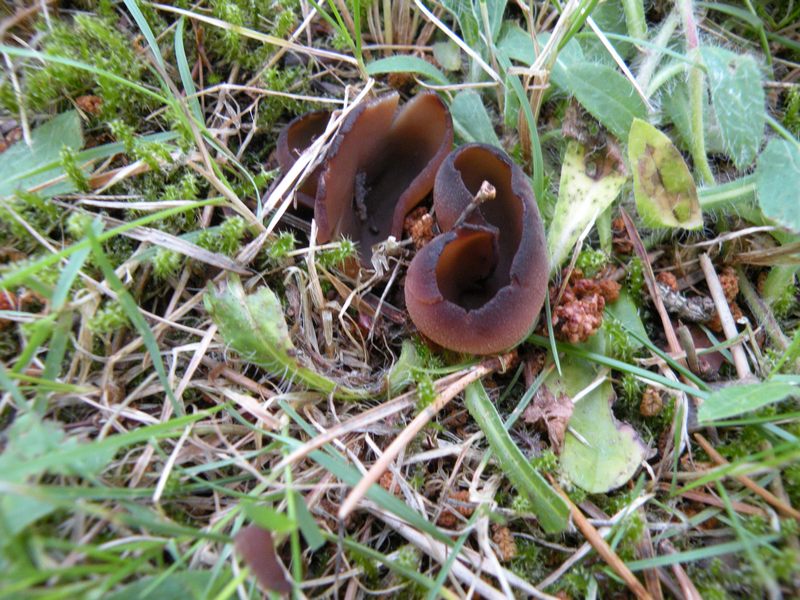}
\includegraphics[height=1.19cm]{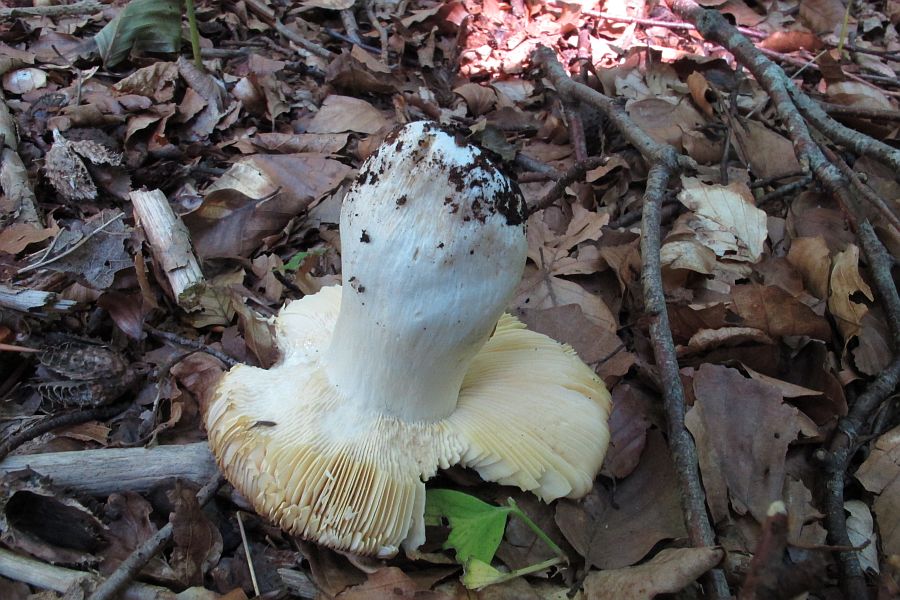}
\includegraphics[height=1.19cm]{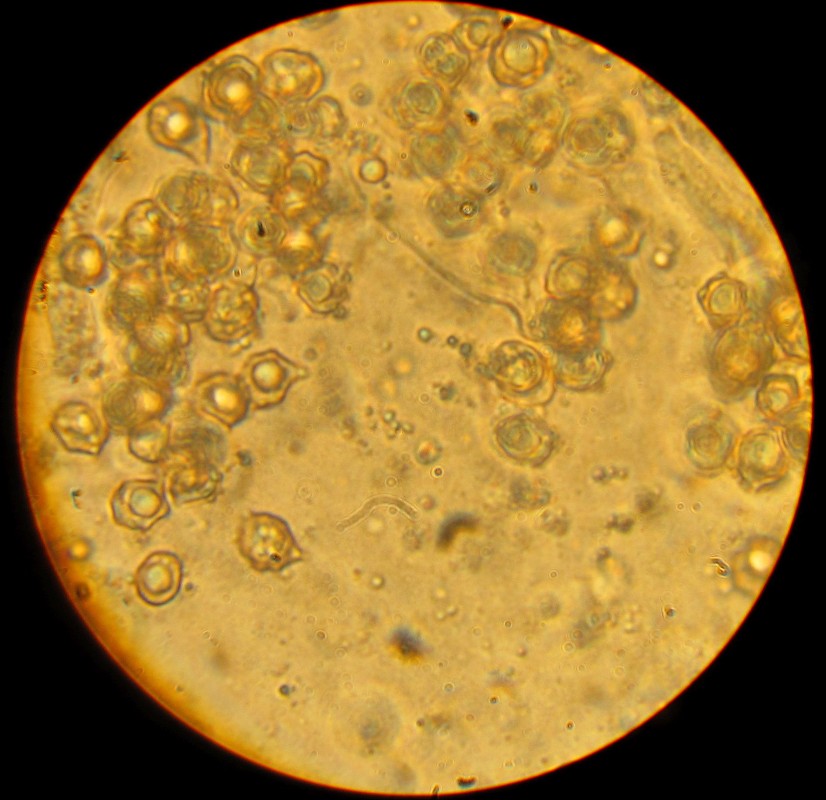}
\\
\includegraphics[height=1.1cm]{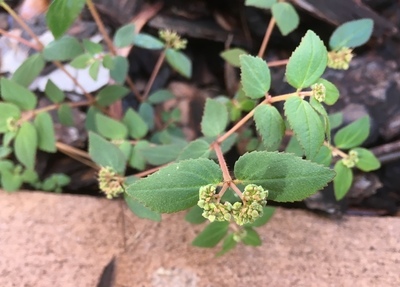}
\includegraphics[height=1.1cm]{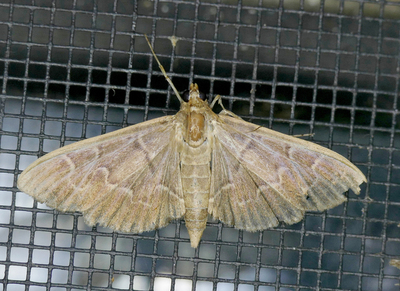}
\includegraphics[height=1.1cm]{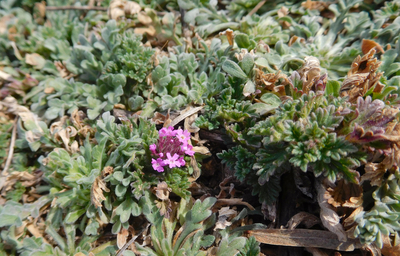}
\includegraphics[height=1.1cm]{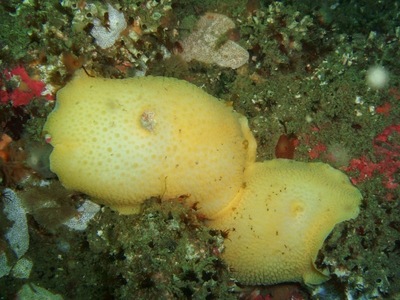}
\includegraphics[height=1.1cm]{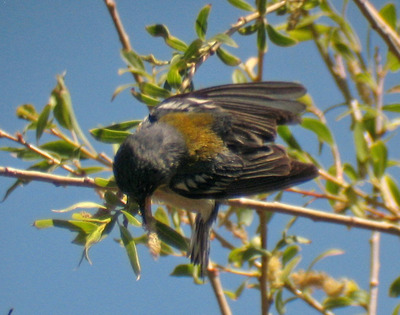} \\
\includegraphics[height=1.1cm]{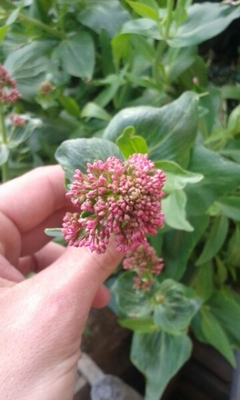}
\includegraphics[height=1.1cm]{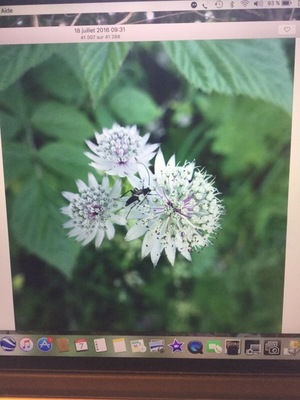}
\includegraphics[height=1.1cm]{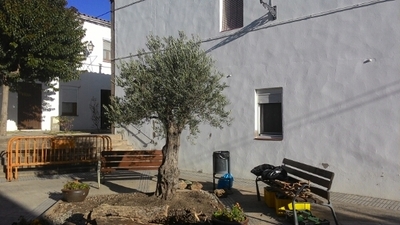}
\includegraphics[height=1.1cm]{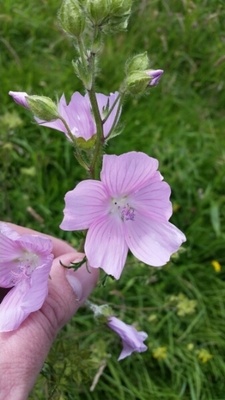}
\includegraphics[height=1.1cm]{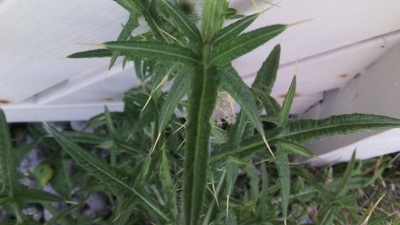}
\includegraphics[height=1.1cm]{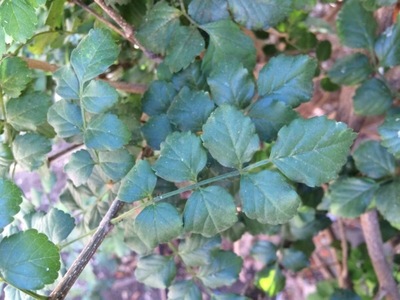}
\\
\caption{Examples from the fine-grained species datasets FGVCx Fungi 2018 (top row), FGVC iNaturalist 2018 (middle row), and PlantCLEF 2017 (bottom row).}
\label{fig:datasets_species}
\end{figure}

Problems related to the differences between training- and test-set data distributions are studied in the field of domain adaptation. We are, however, interested in the special case where statistical properties of observations from the same class stay the same (i.e. appearance does not change), and the only assumed difference is in the class priors~$p(c_k)$.

Methods \cite{du2014semi, saerens2002adjusting} for adjusting classifier outputs to new and unknown a-priori probabilities on the test set have been published years ago, yet the problem of changed class priors is commonly not addressed in computer vision tasks where the situation arises. An exception is the work of Royer et Lampert \cite{royer2015classifier}, who consider the case of sequential adaptation at prediction time (i.e. sample after sample) and take a classical Bayesian approach, using a symmetric Dirichlet distribution as prior information to form a posterior (mean) predictive estimate.

This paper focuses mainly on the case where a whole dataset is available at test time. Adopting the Maximum Likelihood Estimation (MLE) approach of Saerens et al. \cite{saerens2002adjusting} and Du Plessis et Sugiyama \cite{du2014semi}, we propose an alternative solver for the MLE optimization, and we formulate a new Maximum a Posteriori (MAP) estimation approach introducing a Dirichlet hyperprior to increase the stability of the estimator.

We highlight the importance of expecting and adapting to the change of class priors, and we show that such practices can lead to state-of-the-art results in fine-grained visual classification. While our experiments focus on Neural Networks, the proposed framework is applicable to all classifier with probabilistic (posterior) outputs.

Section \ref{section:formulation} provides a  formulation of the problem: a probabilistic interpretation of CNN classifier outputs in Section \ref{subsection:CNNoutputs}, compensation for the change in a-priori class probabilities in Section \ref{subsection:newpriors} and estimation of the new a-priori probabilities using the frameworks of Maximum Likelihood in Section \ref{subsection:priorestimationML} and Maximum a Posteriori in Section \ref{subsection:priorestimationMAP}.

Experiments in Section \ref{section:experiments} show that state-of-the-art Convolutional Neural Networks on fine-grained image classification tasks noticeably benefit from the adaptation to new class prior probabilities, and that the Dirichlet hyper-prior introduced to the proposed MAP approach improves the results over the ML estimate on most datasets.

\section{Problem Formulation and Methodology}
\label{section:formulation}

\subsection{Probabilistic interpretation of CNN outputs}
\label{subsection:CNNoutputs}
Let us assume that a Convolutional Neural Network classifier is trained to provide an estimate of posterior probabilities of classes $c_1, \ldots, c_K \in C$ given an image observation $x_i \in X$: 

\begin{equation}
f_\text{CNN}(c_k \vert \mathbf{x}_i, \mathbf{\theta^*}) \approx p(c_k \vert \mathbf{x}_i),
\end{equation}
where $\mathbf{\theta^*}$ are parameters of the trained CNN.

This is a common interpretation of the process of training a deep network by minimizing the cross-entropy loss $L_\text{CE}$ over samples $\mathbf{x}_i$ with known class-membership labels $c_{ik}$:

\begin{equation}
\begin{aligned} 
&\mathbf{\theta}^* = \argmin_\mathbf{\theta} L_\text{CE} = \argmin_\mathbf{\theta} - \sum\limits_{i=1}^N \sum\limits_{k=1}^K c_{ik} \log f(c_k \vert \mathbf{x}_i, \theta)\\
&= \argmax_\mathbf{\theta} \sum\limits_{i=1}^N \log f(c_{y_i} \vert \mathbf{x}_i, \mathbf{\theta}) = \argmax_\mathbf{\theta} \prod\limits_{i=1}^N f(c_{y_i} \vert \mathbf{x}_i, \mathbf{\theta})\\
\label{eq:ce}
\end{aligned} 
\end{equation}

where $c_{ik}$ is a one-hot encoding of class label $y_i$:

\begin{equation}
c_{ik} = \begin{cases}
1 \; \text{if} \; k=y_i \\
0 \; \text{otherwise}
\end{cases}
\end{equation}

\subsection{New a-priori class distribution}
\label{subsection:newpriors}
When the prior class probabilities $p_e(c_k)$ in our validation/test\footnote{We use index $e$ (for evaluation) to denote all evaluation-time distributions.} set differ from the training set, the posterior $ p_e(c_k \vert \mathbf{x}_i) $ changes too.
The probability density function $p(\mathbf{x}_i \vert c_k)$, describing the statistical properties of observations $\mathbf{x}_i$ of class $c_k$, remains unchanged:
\begin{equation}
p(\mathbf{x}_i \vert c_k) = \frac{ p( c_k \vert \mathbf{x}_i)  p(\mathbf{x}_i) }{ p(c_k) }
= p_e(\mathbf{x}_i \vert c_k) = \frac{ p_e( c_k \vert \mathbf{x}_i)  p_e(\mathbf{x}_i) }{ p_e(c_k) }
\end{equation}

Since $\sum\limits_{k=1}^{K} p_e( c_k \vert \mathbf{x}_i) = 1$ , we can get rid of the unknown probabilities $p(\mathbf{x}_i),p_e(\mathbf{x}_i)$ of fixed sample $\mathbf{x}_i$:

\begin{equation}
\begin{aligned} 
p_e( c_k \vert \mathbf{x}_i) & = p( c_k \vert \mathbf{x}_i) \frac{p_e(c_k) p(\mathbf{x}_i) }{p(c_k) p_e(\mathbf{x}_i)} = \\
& = 
\dfrac{ p( c_k \vert \mathbf{x}_i) \dfrac{p_e(c_k) }{p(c_k)}}{ \sum\limits_{j=1}^{K} p( c_j \vert \mathbf{x}_i) \dfrac{p_e(c_j) }{p(c_j)} } 
\propto
p( c_k \vert \mathbf{x}_i) \dfrac{p_e(c_k) }{p(c_k)}
\end{aligned} 
\label{eq:correction}
\end{equation}

The class priors $p(c_k)$ can be empirically quantified as the number of images labeled as $c_k$ in the training set.
The test-time priors $p_e(c_k)$ are, however, often unknown at test time.

\subsection{ML estimate of new a-priori probabilities}

\label{subsection:priorestimationML}
Saerens et al. \cite{saerens2002adjusting} proposed to approach the estimation of unknown test-time a-priori probabilities by iteratively maximizing the likelihood of the test observations:

\begin{equation}
\begin{aligned} 
L(\mathbf{x}_1,...\mathbf{x}_N)  & = \prod\limits_{i=1}^{N} p_e(\mathbf{x}_i) = \prod\limits_{i=1}^{N} \left[ \sum\limits_{k=1}^{K}   p_e(\mathbf{x}_i, c_k) \right] = \\
& = \prod\limits_{i=1}^{N} \left[ \sum\limits_{k=1}^{K}  p(\mathbf{x}_i \vert c_k) p_e(c_k) \right]
\end{aligned} 
\end{equation}

They derive a simple EM algorithm comprising of the following steps:

\begin{equation}
p_e^{(s)}(c_k \vert \mathbf{x}_i) = \dfrac{ p( c_k \vert \mathbf{x}_i) \dfrac{p_e^{(s)}(c_k) }{p(c_k)}}{ \sum\limits_{j=1}^{K} p( c_j \vert \mathbf{x}_i) \dfrac{p_e^{(s)}(c_j) }{p(c_j)} }
\label{eq:e-step}
\end{equation}
\begin{equation}
p_e^{(s+1)} (c_k) = \frac{1}{N} \sum\limits_{i=1}^{N} p_e^{(s)} (c_k \vert \mathbf{x}_i)
\label{eq:m-step}
\end{equation}
where Eq. \ref{eq:e-step} is the Expectation-step, Eq. \ref{eq:m-step} is the Maximization-step, and $p_e^0(c_k)$  may be initialized, for example, by the training set relative frequency $\approx p(c_k)$.

Du Plessis and Sugiyama \cite{du2014semi} proved that this procedure is equivalent to fixed-point-iteration optimization of the Kullback-Leibler divergence minimization between the test observation density $p_e(\mathbf{x})$ and a linear combination of the class-wise predictions $ q_e(\mathbf{x}) = \sum\limits_{k=1}^{K} P_k p(\mathbf{x}|c_k)$, where $P_k$ are the estimates of $p_e(c_k)$.

\begin{equation}
\begin{aligned}
& \text{KL}(q_e \Vert p_e) = \int p_e(\mathbf{x}) \log\dfrac{p_e(\mathbf{x})}{q_e(\mathbf{x})} d\mathbf{x}
= \\
& = \int p_e(\mathbf{x}) \log p_e(\mathbf{x}) d\mathbf{x} - \int p_e(\mathbf{x}) \log \sum\limits_{k=1}^{K} P_k p(\mathbf{x}|c_k)  d\mathbf{x} 
\end{aligned}
\end{equation}

Note that estimating the priors $\mathbf{P}^\text{MLE} = (P_1,.., P_K)$ by minimization of the KL divergence on the test set $(\mathbf{x}_1,..,\mathbf{x_N})$ can be rewritten as maximization of the log-likelihood $\ell (\mathbf{x}_1,..,\mathbf{x}_N) = \log L (\mathbf{x}_1,..,\mathbf{x}_N) $ of the observed data given the prior probability estimates $P_k \approx p_e(c_k)$:
\begin{equation}
\hspace*{-4mm}
\begin{aligned}
& \argmin\limits_{\mathbf{P}} \text{KL}(q_e \Vert p_e) = \argmax\limits_{\mathbf{P}} \dfrac{1}{N}
\underbrace{\sum\limits_{i=1}^{N} \log \sum\limits_{k=1}^{K}  P_k p(\mathbf{x}_i|c_k)}_{\ell} \\
& \text{s.t. } \sum\limits_{k=1}^{K} P_k = 1 ; \; \forall k : P_k \geq 0
\end{aligned}
\label{eq:optimization-problem1}
\end{equation}

Equation \ref{eq:optimization-problem1} can be further developed into:
\begin{equation}
\begin{aligned}
\mathbf{P}^\text{MLE} & = \argmax\limits_{\mathbf{P}} \sum\limits_{i=1}^{N} \log \sum\limits_{k=1}^{K}  P_k \dfrac{p(c_k | \mathbf{x}_i) p(\mathbf{x}_i)}{p(c_k)} \\
& = \argmax\limits_{\mathbf{P}} \sum\limits_{i=1}^{N}  \log \sum\limits_{k=1}^{K}  P_k 
\underbrace{ \dfrac{p(c_k | \mathbf{x}_i)}{p(c_k)} }_{a_{ik}}
 \\
& \text{s.t. } \sum\limits_{k=1}^{K} P_k = 1 ; \; \forall k : P_k \geq 0
\label{eq:optimization-problem}
\end{aligned}
\end{equation}

As Du Plessis and Sugiyama \cite{du2014semi} have shown, using the EM algorithm from Eq. \ref{eq:e-step}, \ref{eq:m-step} may not result in the unique optimal value, as the mapping of the fixed-point iteration is not a contraction mapping.

We therefore experiment also with direct optimization of the objective from Eq. \ref{eq:optimization-problem} using the projected gradient descent algorithm \cite{boyd2004convex}, or more precisely projected gradient ascent if we consider the maximization task. At each step $s$, we update the variables as follows:

\begin{equation}
    P_k^{(s+1)} = \pi \left( P_k^{(s)} + \lambda \dfrac{\partial  \ell(\mathbf{x}_1,..,\mathbf{x}_N) }{\partial P_k} \right),
\end{equation}

where $\lambda$ is the learning rate, $\pi$ represents projection onto the unit simplex (i.e. on the constraint set from  Eq. \ref{eq:optimization-problem}) and the partial derivatives are:

\begin{equation} \dfrac{\partial  \ell(\mathbf{x}_1,..,\mathbf{x}_N) }{\partial P_k} = \sum\limits_{i=1}^{N} \dfrac{a_{ik}}{ \sum\limits_{j=1}^K P_j a_{ij} }
\end{equation}

To compute the Euclidean projection $\pi$ onto the unit simplex, we use the efficient algorithm from \cite{duchi2008efficient, wang2013projection}.

\subsection{MAP estimate of new a-priori probabilities}
\label{subsection:priorestimationMAP}
Having a prior knowledge on the categorical distribution, $p(\mathbf{P})$, the maximum a-posteriori (MAP) estimate of the class prior probabilities is:
\begin{equation}
\begin{aligned}
\mathbf{P}^\text{MAP} & = \argmax\limits_{\mathbf{P}} p(\mathbf{P} \vert (\mathbf{x}_1,..,\mathbf{x}_N)) \\
& = \argmax\limits_\mathbf{P} p(\mathbf{P}) \prod\limits_{i=1}^N p(\mathbf{x}_i \vert \mathbf{P}) \\
& = \argmax\limits_{\mathbf{P}} \log p(\mathbf{P}) + \sum\limits_{i=1}^N \log p(\mathbf{x}_i \vert \mathbf{P}) \\
& \text{s.t. } \sum\limits_{k=1}^{K} P_k = 1 ; \; \forall k : P_k \geq 0
\end{aligned}
\label{eq:map-optimization}
\end{equation}

Note that the second term is the log-likelihood from the previous section, $\ell(\mathbf{x}_1,..,\mathbf{x}_N) = \sum\limits_{i=1}^N \log p(\mathbf{x}_i \vert \mathbf{P})$.

Let us model the prior knowledge about the categorical distribution by the symmetric Dirichlet distribution:

\begin{equation}
    p(\mathbf{P}) = \dfrac{1}{B(\alpha)} \prod\limits_{k=1}^{K} P_k^{\alpha-1}
\end{equation}

parametrized by $\alpha > 0$, where the normalization factor for the symmetric case is $B(\alpha) = \dfrac{\Gamma(\alpha)^K}{\Gamma(\alpha K)}$.

Choosing an $\alpha \geq 1$ favours dense distributions, and thus avoids setting the categorical priors too close to zero. Zero priors may suppress even highly confident predictions. Moreover, the Dirichlet distribution with $\alpha\geq1$ is a log-concave distribution, which is suitable for the optimization of Eq. \ref{eq:map-optimization}.
    
The optimization for $\alpha \geq 1$ can be performed by the projected gradient descent optimizer from Section \ref{subsection:priorestimationML} by adding the following gradient components:
\begin{equation}
\dfrac{\partial \log p(\mathbf{P}) }{\partial P_k} = \dfrac{\partial (\alpha - 1)\log(P_k)}{\partial P_k} = \dfrac{\alpha - 1}{P_k}
\end{equation}

\section{Experiments}
\label{section:experiments}

\begin{figure*}
\makebox[\textwidth][c]{
\includegraphics[width=0.9\textwidth]{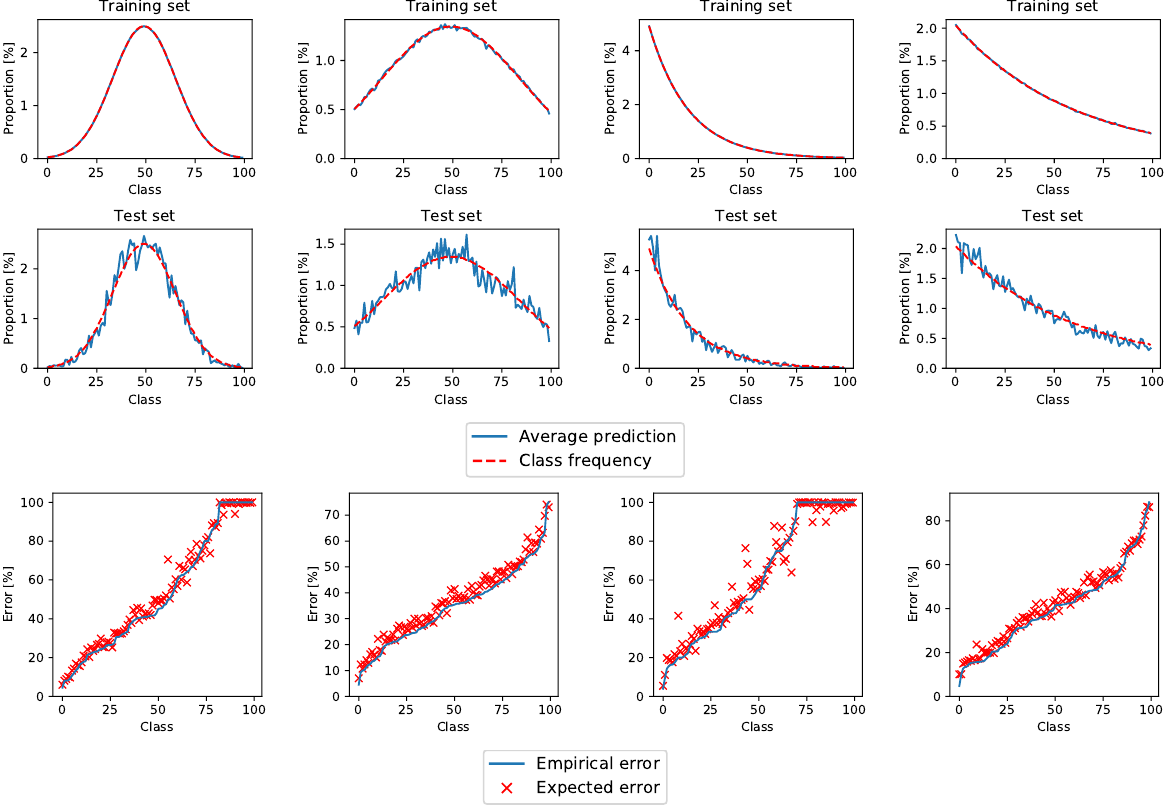}
}
\caption{Top and middle row: Comparison of class frequency and CNN output marginalization over all images in the train- and test- sets sampled from CIFAR-100. Bottom row: Comparison of the test set empirical error $ \epsilon^{\text{emp}}_k $ and the expected error $ \epsilon_k$, sorted by $ \epsilon^{\text{emp}}_k $.}
\label{fig:marginalization}
\end{figure*}

The following fine-grained classification datasets are used for experiments in this Section:

\textbf{CIFAR-100} is a popular dataset for smaller-scale fine-grained classification experiments, introduced by Krizhevsky  and Hinton \cite{krizhevsky2009learning} in 2009. It contains small resolution (32x32) color images of 100 classes. While the dataset is balanced (with 500 training samples and 100 test samples for each class), we sample a number of its unbalanced subsets for our experiments in this Section.

\textbf{PlantCLEF 2017} \cite{goeau2017plant} was a plant species recognition challenge organized as part of the LifeCLEF workshop \cite{joly2017lifeclef}.
The provided training images for 10,000 plant species consisted from an EOL "trusted" training set (downloaded from the Encyclopedia of Life\footnote{\url{http://www.eol.org/}}), a significantly larger "noisy" training set (obtained from Google and Bing image search results, including mislabeled or irrelevant images), and the previous years (2015-2016) images depicting only a subset of the species.
We use the training data in two ways: Either training on all the sets together (including the "noisy" set) - further denoted as \textit{PlantCLEF-All}, or excluding the "noisy" set (i.e. using the 2017 EOL data and the previous years data) - further denoted as \textit{PlantCLEF-Trusted}.
The test set from the PlantCLEF 2017 challenge is used for evaluation.
All data is publicly available\footnote{\url{http://www.imageclef.org/lifeclef/2017/plant}, \url{http://www.imageclef.org/node/198}}. PlantCLEF presents an example of a real-world fine-grained classification task, where the number of available images per class is highly unbalanced.

\textbf{FGVC iNaturalist 2018} is a large scale species classification competition, organized with the FGVC5 workshop at CVPR 2018. The provided dataset covers 8,142 species of plants, animals and fungi. The training set is highly unbalanced and contains almost 440K images. A balanced validation set of 24K images is provided.

\begin{figure*}
\begin{center}\begin{tabular}{lll}
\includegraphics[height=2.01cm]{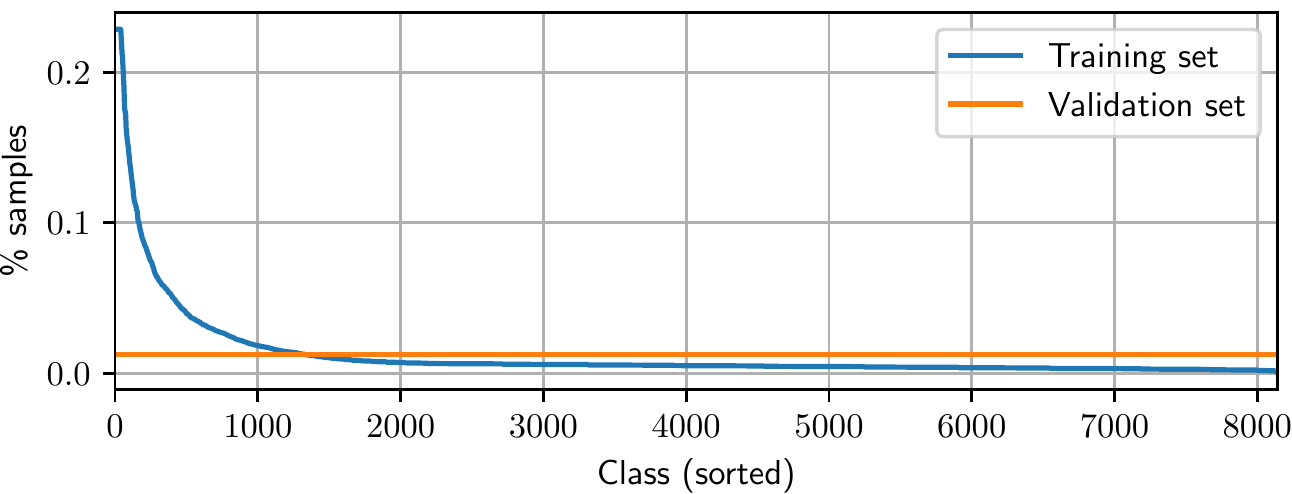} &
\includegraphics[height=2.01cm]{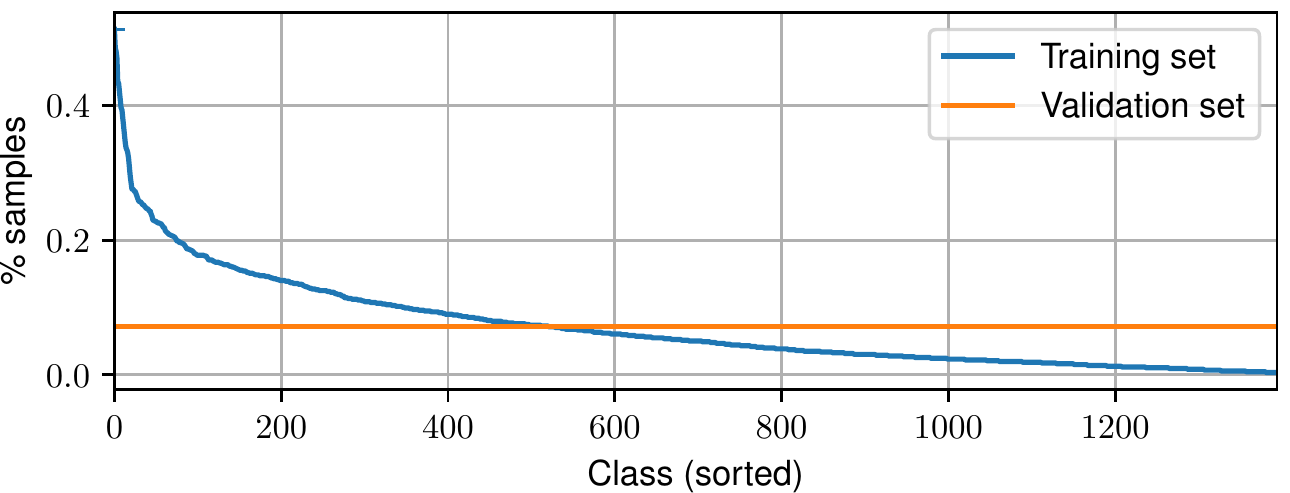} &
\includegraphics[height=2.01cm]{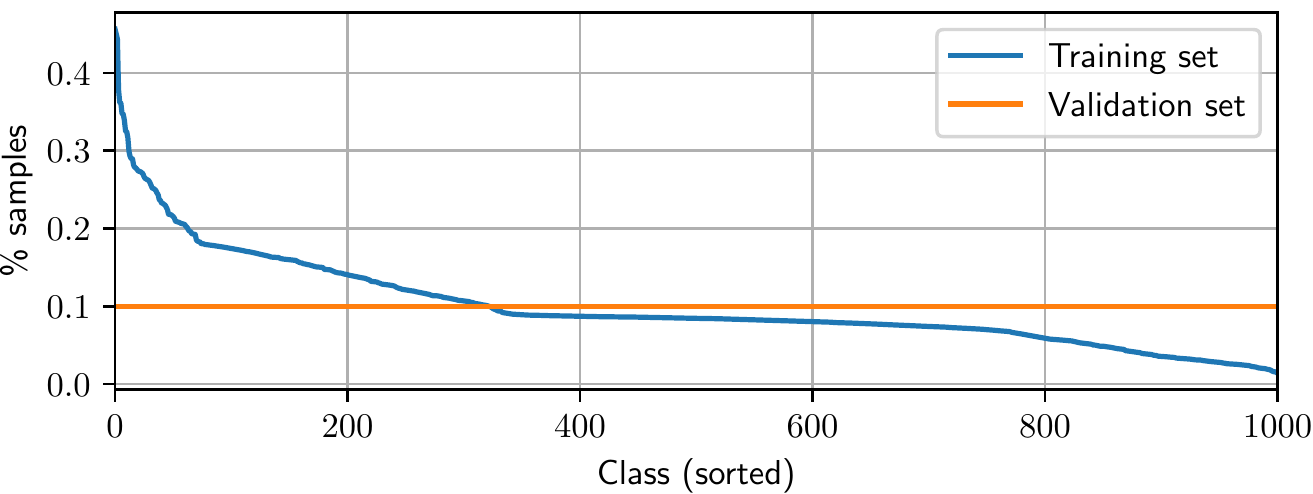} \\
\hspace*{4pt}\includegraphics[height=3.5cm]{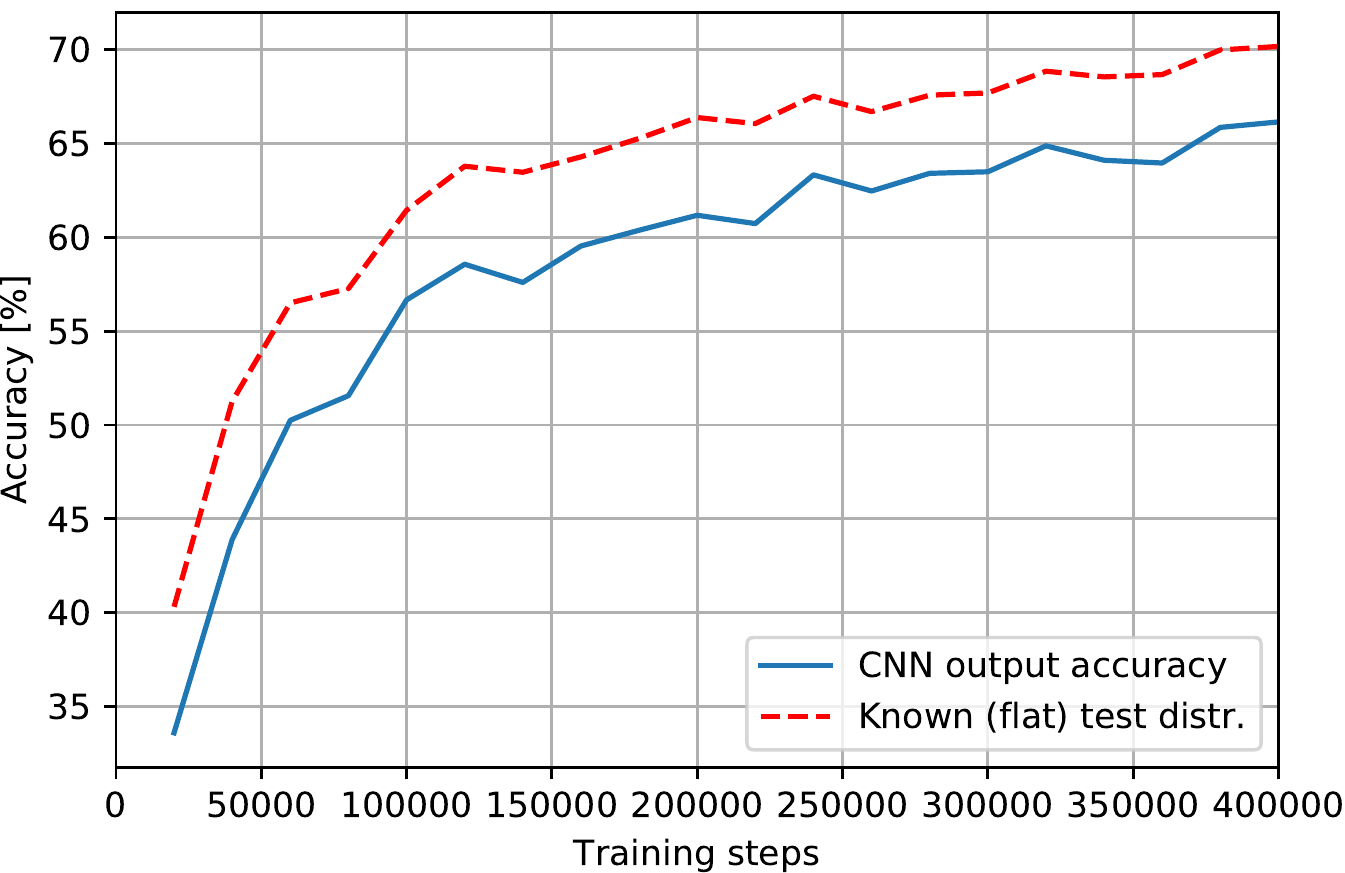} &
\hspace*{-2pt}\includegraphics[height=3.5cm]{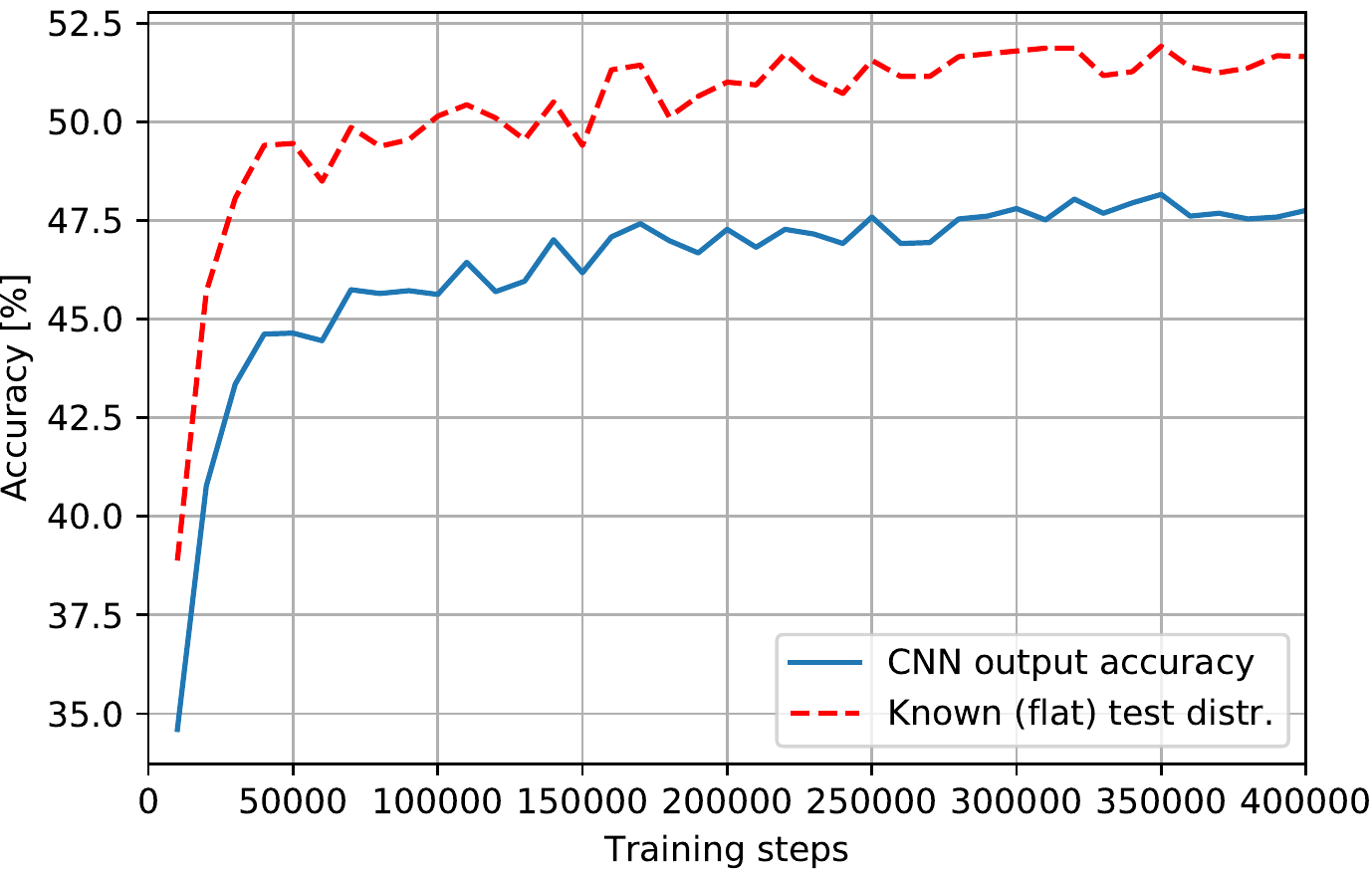} &
\hspace*{-2pt}\includegraphics[height=3.5cm]{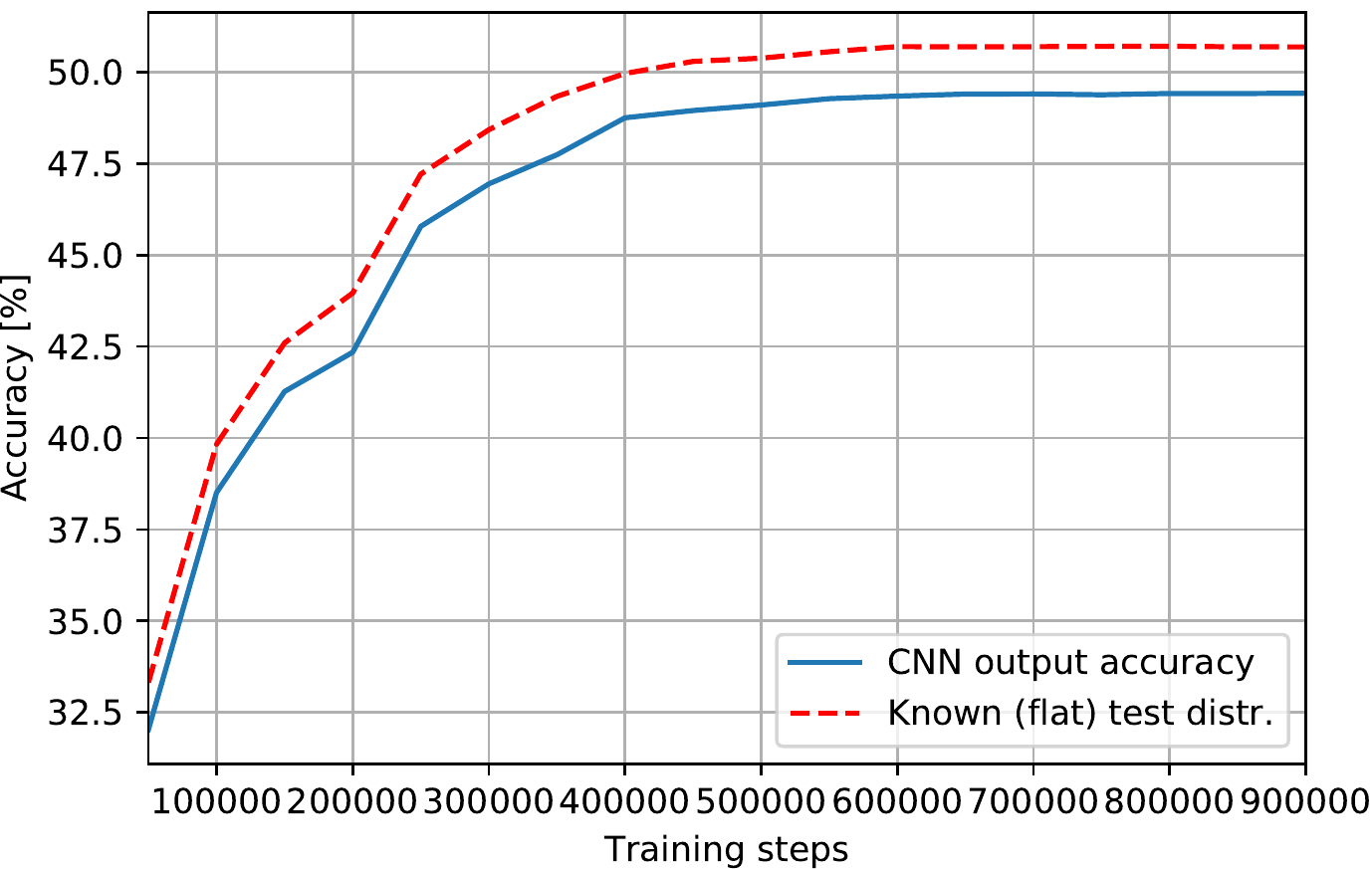}
\end{tabular}\end{center}
\caption{Training and validation set distributions (top) and accuracy before and after correcting predictions with the known/uniform val. set distribution (bottom) for FGVC iNaturalist 2018 (left), FGVCx Fungi 2018 (middle) and Webvision 2017 (right).}
\label{fig:fgvc}
\end{figure*}

\textbf{FGVCx Fungi 2018} is a another species classification competition, focused only on fungi, also organized with the FGVC5 workshop at CVPR 2018. The dataset covers nearly 1,400 fungi species. The training set contains almost 86K images, and is highly unbalanced. The validation set is balanced, with 4,182 images in total.

\textbf{Webvision 1.0} \cite{li2017webvision} (also Webvision 2017) is a large dataset designed to facilitate learning visual representation from noisy web data. It contains more than 2.4 million of images crawled from Flickr and Google Images and covers the same 1,000 classes as the ILSVRC 2012 dataset. The number of images per category ranges from hundreds to more than 10 thousand, depending on the number of queries generated from the synset for each category and on the availability of images on the Flickr and Google.

Examples from the FGVC and PlantCLEF datasets are displayed in Figure \ref{fig:datasets_species}.

\subsection{Validation of posterior estimates on the training set}
\label{section:validation_posteriors}

Before considering the change in class priors, let us validate that the marginalization of CNN predictions on training and validation data estimates the class priors well:
\begin{equation}
p(c_k) = \frac{1}{N} \sum\limits_{i=1}^{N} p (c_k \vert \mathbf{x}_i)  \approx \frac{1}{N} \sum\limits_{i=1}^{N} f_\text{CNN} (c_k \vert \mathbf{x}_i) \approx \dfrac{N_k}{N},
\end{equation}
where $N_k = \sum\limits_{i=1}^{N} c_{ik} $  is the number of images of class $c_k$.
We simulated normal and exponential prior class distributions by randomly picking subsets of the CIFAR-100 database that follow the chosen distributions. A 32-layer Residual Network\footnote{Implementation from \url{https://github.com/tensorflow/models/tree/master/research/resnet}} \cite{he2016deep} was trained on the training-subsets.
The comparison of empirical class frequencies and the estimates obtained by marginalizing the CNN outputs (i.e. averaging CNN predictions) is displayed in the upper part of Figure \ref{fig:marginalization}. The training set class distributions are estimated almost perfectly. The estimates on the test set are more noisy, but still approximate the class frequencies well.

Let us also compare the expected error $ \epsilon_k$ and the empirical error $ \epsilon^{\text{emp}}_k $ for each class $c_k$:
\begin{equation}
\epsilon_k =  \dfrac{1}{N_k} \sum\limits_{i: y_i=k} 1 - p(c_k | x_i) ,
\end{equation}
\begin{equation}
\epsilon^{\text{emp}}_k = \dfrac{1}{N_k} \sum\limits_{i: y_i=k} [\![ k \neq \argmax\limits_{c_j} f_\text{CNN}(c_j \vert \mathbf{x}_i) ]\!] ,
\end{equation}

The comparison of the test set empirical error $ \epsilon^{\text{emp}}_k $ and the expected error  $ \epsilon_k$ , displayed in the bottom part of Figure \ref{fig:marginalization}, also suggest a good estimate of posterior probabilities.

\subsection{Adjusting posterior probabilities when test-time priors are known}
\label{section:prior_known}

To experiment with known test-time prior probabilities $p_e(c_k)$, we use the training and validation sets from the FGVC iNaturalist\footnote{\url{https://sites.google.com/view/fgvc5/competitions/inaturalist}} Competition 2018 and the FGVCx Fungi\footnote{\url{https://sites.google.com/view/fgvc5/competitions/fgvcx/fungi}} Classification Competition 2018. In these challenges, the validation sets are balanced (i.e. the class prior distribution is uniform).
A state-of-the-art Convolutional Neural Network, Inception-v4 \cite{szegedy2017inception}, was fine-tuned for each task. The predictions were corrected as prescribed by Equation \ref{eq:correction}.

A similar case is the Webvision 2017 dataset, where the training set is highlt unbalanced and the validation set is balanced. In the classification/baseline experiments of Li et~al. \cite{li2017webvision}, the change of class prior probabilities is not taken into consideration. Similarly to \cite{li2017webvision} we train an AlexNet network from scratch. (Note that our model did not converge to the same accuracy, probably due to difference in implementation and hyper-parameters.)

Figure \ref{fig:fgvc} displays the training and evaluation distribution and the improvement in accuracy achieved by correcting the predictions with the known priors. The improvement in top-1 accuracy is \textbf{4.0\%} and \textbf{3.9\%} after 400K training steps (and up to \textbf{7.4\%} and \textbf{4.9\%} during fine-tuning) for the FGVC iNaturalist and FGVCx Fungi classification challenges respectively and \textbf{1.3\%} for the Webvision 2017 dataset.

\begin{figure*}[h]
\makebox[\textwidth][c]{
\includegraphics[width=\textwidth]{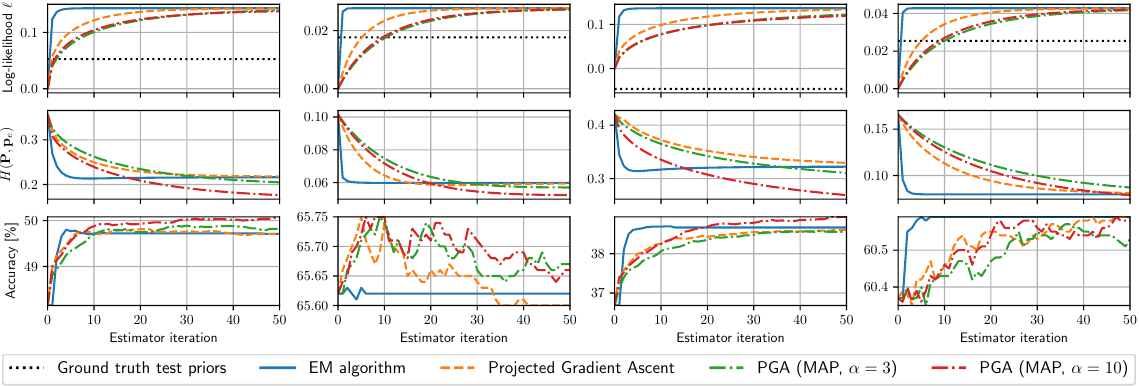}
}
\caption{Iterative estimation of test-time priors on the full CIFAR-100 test set from CNN estimates trained on unbalanced CIFAR-100 subsets (same order as in Figure \ref{fig:marginalization}).}
\label{fig:cifar-estimation}
\end{figure*}

\begin{table*}
\makebox[\textwidth][c]{
\begin{tabular}{l c c c c c c c c c c c c }
Train. distribution & \includegraphics[width=8mm]{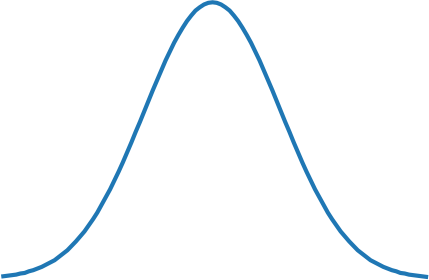} & \includegraphics[width=8mm]{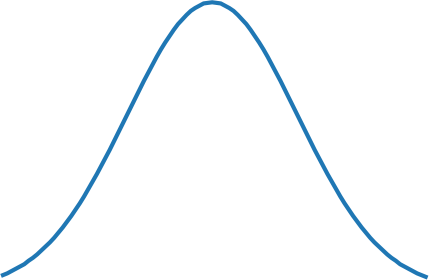} & \includegraphics[width=8mm]{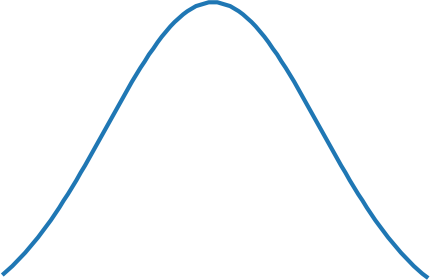} & \includegraphics[width=8mm]{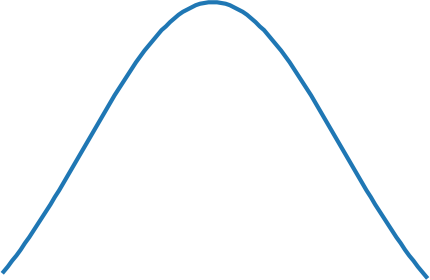} & \includegraphics[width=8mm]{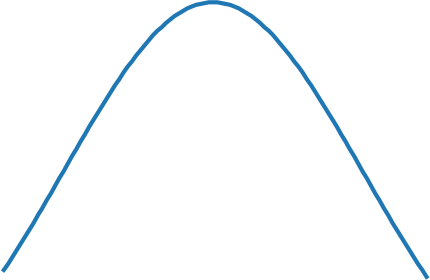} & \includegraphics[width=8mm]{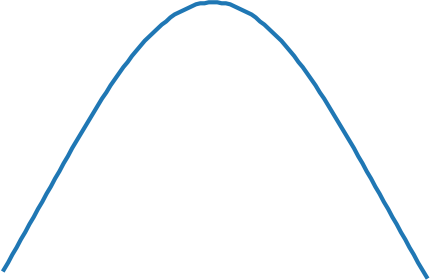} & \includegraphics[width=8mm]{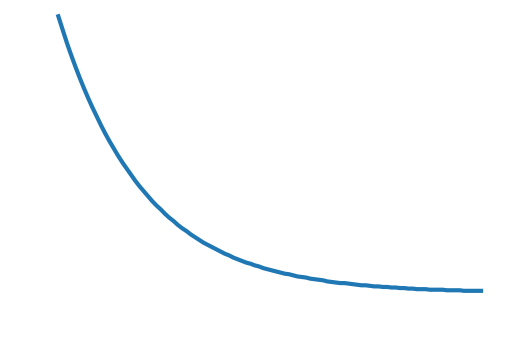} & \includegraphics[width=8mm]{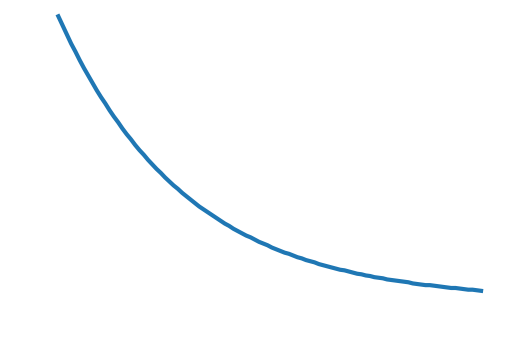} & \includegraphics[width=8mm]{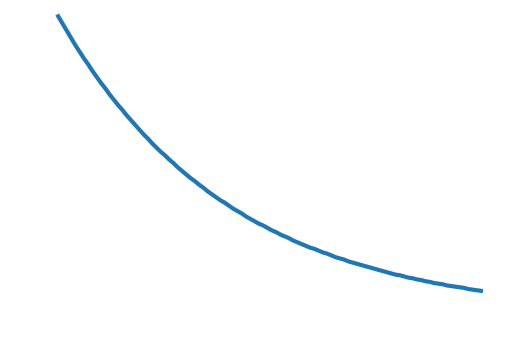} & \includegraphics[width=8mm]{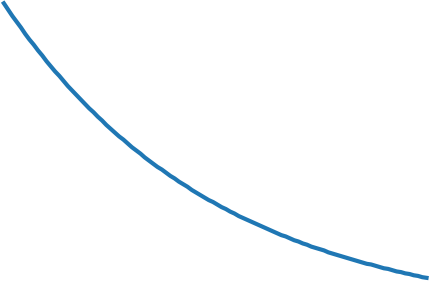} & \includegraphics[width=8mm]{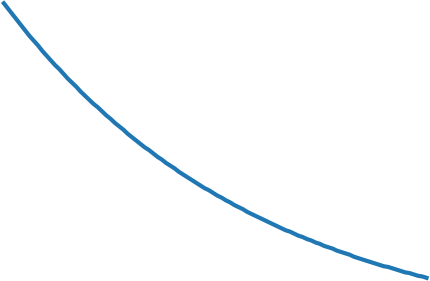} & \includegraphics[width=8mm]{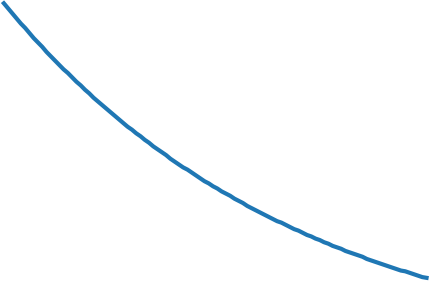} \\ \hline

Acc.[\%] & 48.15 & 55.70 & 60.88 & 64.01 & 65.62 & \textbf{67.29} & 36.68 & 47.72 & 54.00 & 56.57 & 60.37 & 61.66 \\
-- after ML (EM) & 49.71 & 56.94 & 61.64 & 64.58 & 65.62 & 67.11 & 38.67 & 49.05 & 55.18 & 57.05 & 60.59 & 61.74 \\
-- after ML (PGA) & 49.71 & 56.94 & 61.64 & 64.58 & 65.62 & 67.11 & 38.67 & 49.05 & 55.18 & 57.05 & 60.59 & 61.74 \\
-- after MAP, $\alpha=3$ & 49.75 & 56.94 & 61.65 & \textbf{64.59} & 65.64 & 67.18 & 38.75 & 49.20 & 55.19 & \textbf{57.10} & 60.58 & \textbf{61.76} \\
-- after MAP, $\alpha=10$ & \textbf{50.07} & \textbf{56.97} & \textbf{61.68} & 64.55 & \textbf{65.70} & 67.23 & \textbf{39.12} & \textbf{49.34} & \textbf{55.22} & \textbf{57.10} & \textbf{60.69} & \textbf{61.76} \\
\hline
Acc.[\%] known $p_e(c_k)$ & 51.20 & 57.61 & 62.23 & 64.73 & 65.92 & 67.44 & 40.62 & 50.07 & 55.86 & 57.49 & 60.92 & 62.11 \\ \hline
\end{tabular}
}

\caption{
Accuracy of CNN classifiers trained on unbalanced CIFAR-100 subsets (top) and evaluated on the full CIFAR-100 test set, adjusted by estimated class priors using the EM algorithm and the projected gradient ascent (PGA). Predictions adjusted by an oracle knowing the class priors (bottom).}
\label{table:cifar_experiments}
\end{table*}

\subsection{Adjusting posterior probabilities when the whole test set with unknown priors is available at test-time}
\label{section:estimating_priors}

We choose the PlantCLEF 2017 challenge test set as an example of test environment, where no knowledge about the class distribution was available. The training set is highly unbalanced, the test set statistics does not follow the training set statistics and does not even contain examples from all classes.

We used an Inception-V4 model pre-trained on all available training data (\textit{PlantCLEF-All}). The results in Table \ref{table:lifeclef} show, that the top-1 accuracy increases by \textbf{3.4\%} when estimating the test set priors using the EM algorithm of Saerens et al. \cite{saerens2002adjusting} (Eq. \ref{eq:e-step}, \ref{eq:m-step}). To compare with the results of the 2017 challenge, we combine the predictions per specimen observation (the test set contained several images per specimen, linked by ObservationID meta-data) and compute the observation-identification accuracy. Note that after the test set prior-estimation, our single CNN model outperforms the winning submission of PlantCLEF 2017 composed of 12 very deep CNN models (ResNet-152, ResNeXt-101 and GoogLeNet architectures).

\begin{table*}
\centering
\begin{tabular}{b{38mm}b{16mm}b{25mm}b{35mm}b{35mm}}
  Model & Accuracy &  \shortstack[l]{Acc. after EM} & \shortstack[l]{Acc. per observation,\\after EM} & \shortstack[l]{Acc. per observation,\\$p_e(c_k)$ known} \\
 \hline
Inception V4 & 83.3\% & 86.7\% &  \textbf{90.8}\% & 93.7\% \\
\hline
Ensemble of 12 CNNs \cite{lasseck2017image}
(PlantCLEF2017 winner) & -- & -- &  88.5\% & --\\
\hline
\end{tabular}
\caption{Improvement in accuracy after applying the iterative test set prior estimation in the PlantCLEF 2017 plant identification challenge.}
\label{table:lifeclef}
\end{table*}

\begin{figure*}
\makebox[\textwidth][c]{
\includegraphics[width=0.9\textwidth]{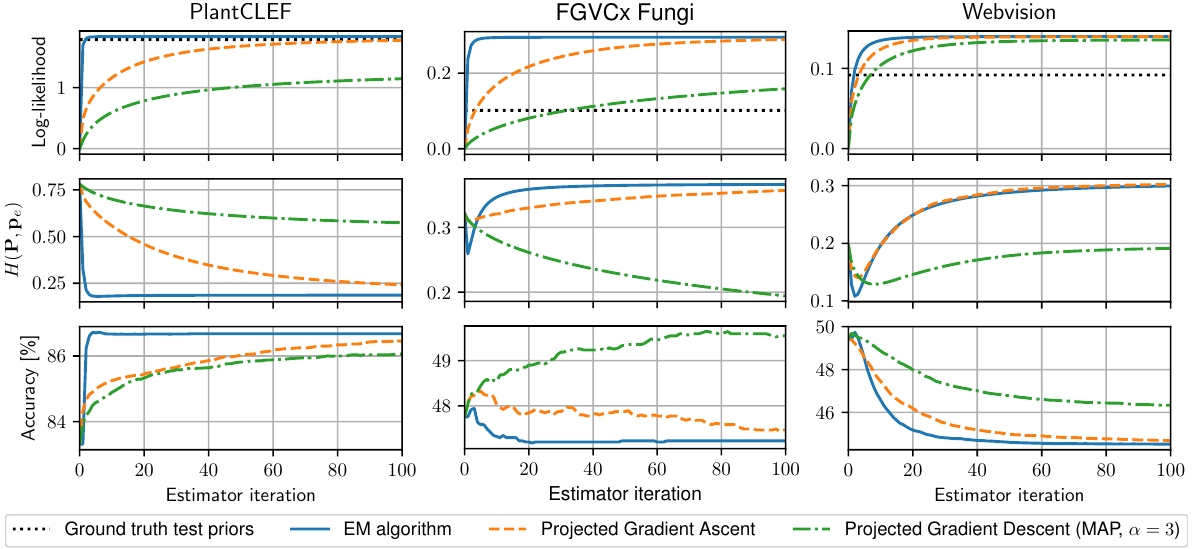}
}
\caption{Iterative estimation of test-time priors on fine-grained datasets: PlantCLEF (Inception-v4), FGVCx Fungi (Inception-v4), and Webvision 1.0 (AlexNet). Top row: The log-likelihood surrogate $\ell$. Middle row: Hellinger distance between the prior estimate and ground truth class frequencies. Bottom row: Accuracy.}
\label{fig:finegrained-estimation}
\end{figure*}

A set of experiments was performed with the networks from Section \ref{section:validation_posteriors} trained on the selected subsets of CIFAR-100. We evaluate the networks against the full (balanced) CIFAR-100 test set, and compare the accuracy of the CNN predictions against the predictions adjusted by estimated priors and predictions adjusted with ground-truth test-time priors. The results are in Table \ref{table:cifar_experiments}. As expected, knowing the ground truth priors would always lead to the best results. With only one exception, estimating the test-time priors always leads to an increase in accuracy. The MAP estimate consistently achieves higher test-time accuracy, although, as illustrated in Figure \ref{fig:cifar-estimation}, the likelihood of its estimate is lower than of the ML estimates. This demonstrates the importance of adding prior assumptions on the estimated class prior probabilities. The EM algorithm for ML estimation, however, converges noticeably faster.

Figure \ref{fig:finegrained-estimation} illustrates the estimation of class priors on the fine grained datasets PlantCLEF, FGVCx Fungi and Webvision. We can notice a positive effect MAP estimation on the FGVCx Fungi dataset, where it increases accuracy by 1.8\%, while ML estimate leads to a decrease in accuracy. On the Webvision dataset, all estimation methods decrease the accuracy, however MAP has the lowest decrease. The poor performance on Webvision may be related to the high amount of outliers in the Webvision training set - Li et al. \cite{li2017webvision} suggest that only 66\% of the images can be considered inliers. This may affect the reliability of the CNN posterior estimate. The accuracy on PlantCLEF increases by 2.8\% after MAP estimation and by 3.4\% after ML estimation. Note that on PlantCLEF, many classes are not present in the test set, and therefore the optimization is actually disadvantaged by the Dirrichlet prior preventing the class prior probabilities from converging to zero.

\subsubsection{Cross-validation of the prior-estimate likelihood on a set without labels}
The experiments in Section \ref{section:estimating_priors} show that increasing the likelihood does not always lead to a more precise estimate. One possible reason may be over-fitting to the predictions on the test set (or to $a_{ik}$ in Equation \ref{eq:optimization-problem}). Let us "cross-validate" the likelihood on the test set: We will optimize the estimate only on a random half of the test set (likelihood-optimization set), and use the other half for likelihood-validation. Note that for this experiment, we use the projected gradient descent with a lower learning rate, in order to observe the changes in convergence in more detail.

\begin{figure}
\makebox[0.48\textwidth][c]{
\includegraphics[width=0.5\textwidth]{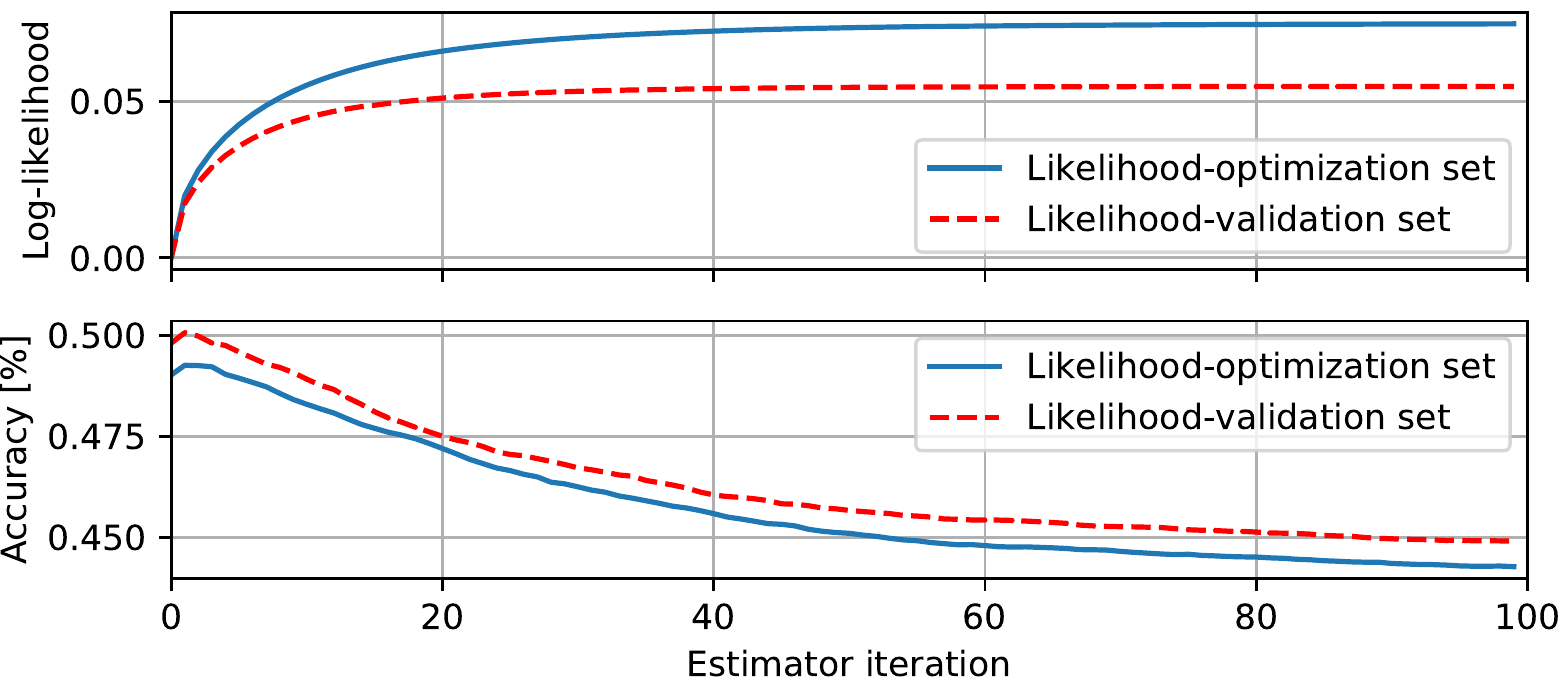}
}
\caption{"Cross-validation" of the likelihood optimization on Webvision 1.0, using only half of the test set (likelihood-optimization set) to estimate the class priors, and observing the log-likelihood on the other half (likelihood-validation set).}
\label{fig:likelihood-crossval}
\end{figure}

Figure \ref{fig:likelihood-crossval} shows, that even for the "unseen" half of the data (likelihood-validation set), the likelihood of the solution still increases, while the accuracy on both sets is decreasing. Therefore, this is not a case of over-fitting to the seen predictions.

\subsection{Adjusting posterior probabilities on-line with new test samples}
\begin{figure}
\centering
\includegraphics[width=0.45\textwidth]{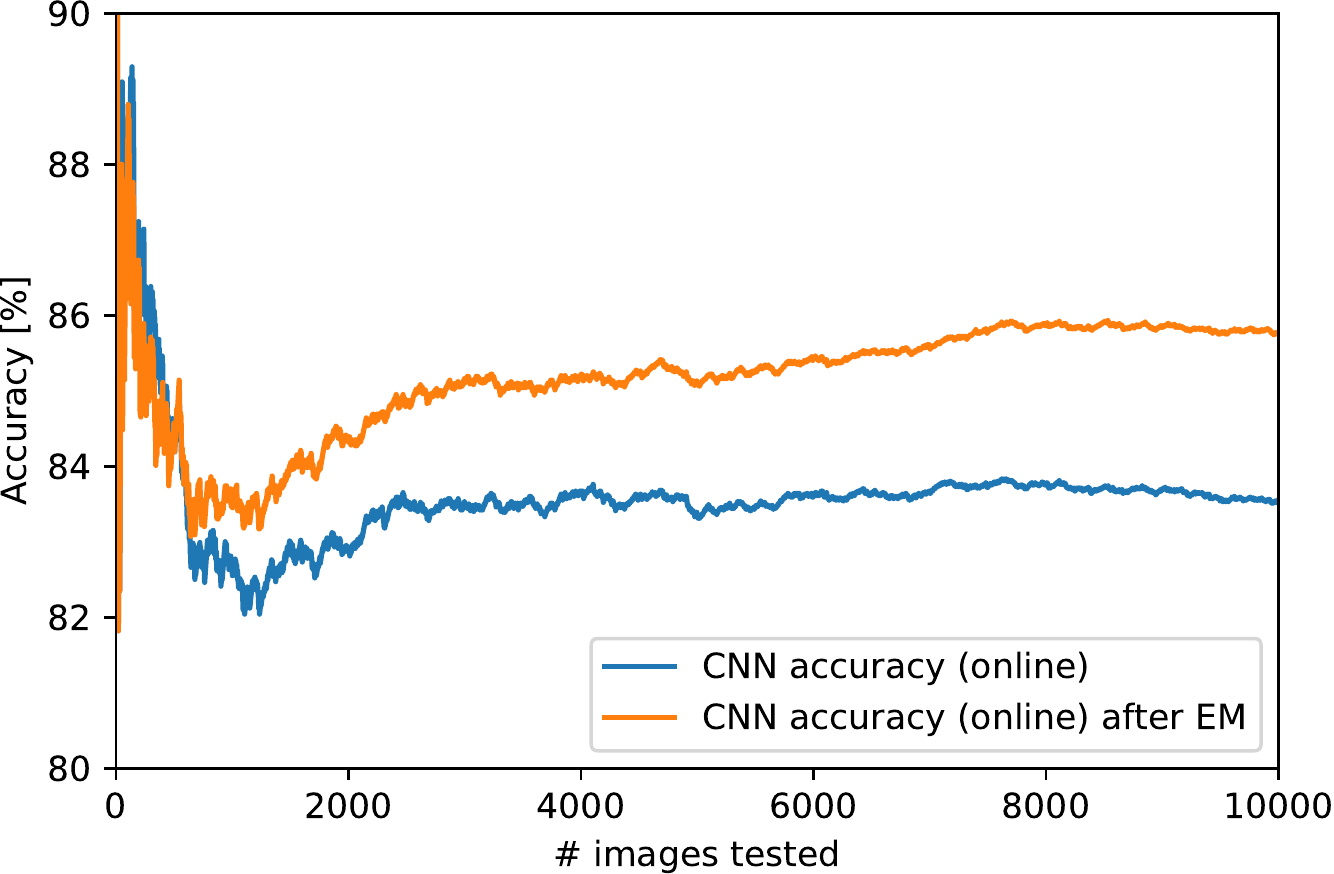}
\caption{On-line test-prior estimation (i.e. images tested sequentially) on the PlantCLEF 2017 dataset.}
\label{fig:plantCLEFonline}
\end{figure}

In practical tasks, test samples are often evaluated rather sequentially than all at once. We evaluate how the test-time class prior estimation on the PlantCLEF 2017 dataset affects the results on-line, i.e. when the priors are always estimated from the already seen examples. In Figure \ref{fig:plantCLEFonline}, after about 1,000 test samples, the predictions adjusted by class priors iteratively estimated by the EM algorithm gain a noticeable margin against the plain CNN predictions. Moreover, the accuracy of the adjusted predictions was not significantly lower than the original predictions even for the first few hundred test cases.

\subsection{Changing the training set priors}

\begin{figure}
\centering
\includegraphics[height=5.3cm]{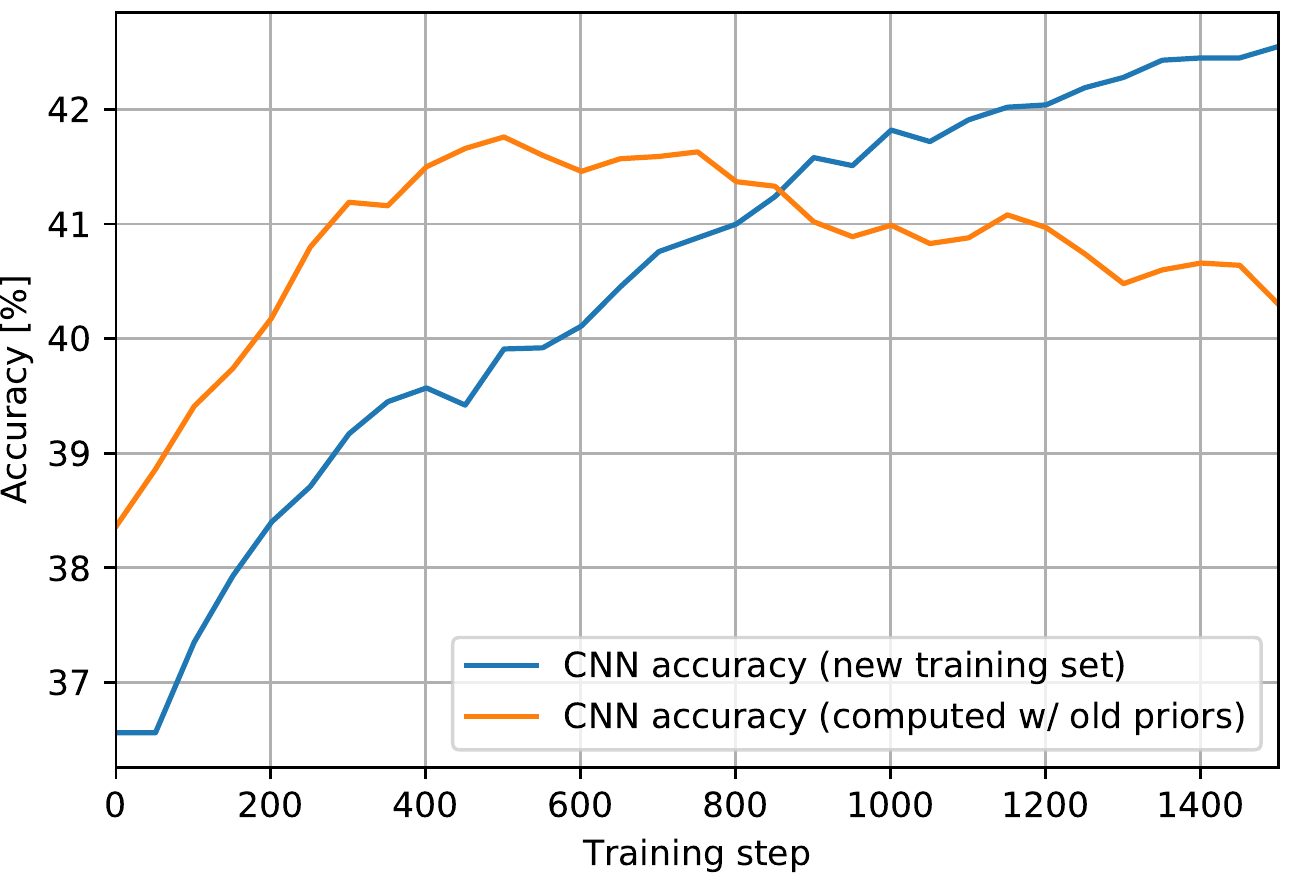}
\caption{CNN pre-trained on unbalanced CIFAR-100 subset fine-tuned on the full CIFAR-100 training set.} 
\label{fig:trainchange}
\end{figure}

How fast does the effective "learned" priors change when the training set changes during training? In this experiment, new samples are added into the training set. We take a network from Section \ref{section:validation_posteriors} pre-trained on an unbalanced subset of CIFAR-100 and we fine-tune it on the full (balanced) CIFAR-100 training set. The predictions are evaluated on the complete (and balanced) test set.  From the results in Figure \ref{fig:trainchange}, it is clearly visible that using the old training set priors is still favorable for a few fine-tuning steps, but the effective priors of the CNN classifier seem to change fast.

\section{Conclusions}

The paper highlighted the importance of not ignoring the commonly found difference between the class priors in the training and test sets in computer vision.
We compared two approaches to estimating the test set priors: the existing Maximum Likelihood Estimation approach (maximizing the test observation likelihood by an existing EM-based method \cite{saerens2002adjusting}  algorithm and by projected gradient ascent) and the proposed Maximum a Posteriori approach, putting the Dirichlet prior on the categorical distributions.

Experimental results show a significant improvement on the FGVC iNaturalist 2018 and FGVCx Fungi 2018 classification tasks using the known evaluation-time priors, increasing the top-1 accuracy by 4.0\%  and 3.9\% respectively. Iterative EM estimation of test-time priors on the PlantCLEF 2017 dataset increases the image classification accuracy by 3.4\%, allowing a single CNN model to achieve state-of-the-art results and outperform the competition-winning ensemble of 12 CNNs.
Adding the Dirichlet prior, preventing the class prior estimates from getting too close to zero, brings a slightly lower 2.8\% increase in accuracy on the PlantCLEF dataset (where many classes are actually missing in the test set), but improves the results and stability in most cases, including the FGVCx Fungi dataset, where it increased accuracy by 1.8\% while the ML estimate would lead to a decrease. The only experimental case where the estimate of the unknown prior probability doesn't help is the Webvision dataset - this may be related to the high amount of outliers (Li et al. \cite{li2017webvision} suggest that only 66\% of the images can be considered inliers).

{\small
\bibliographystyle{ieee}
\bibliography{bibliography}
}

\end{document}